\documentclass[
]{ceurart}

\usepackage{fancyvrb}
\usepackage{amssymb}
\usepackage{listings}
\usepackage{graphicx}
\usepackage[export]{adjustbox}
\usepackage[relative,overlay]{textpos}
\usepackage{tikz}
\usepackage{todonotes}
\begin{document}

\copyrightyear{2023}
\copyrightclause{Copyright for this paper by its authors.
  Use permitted under Creative Commons License Attribution 4.0
  International (CC BY 4.0).}

\conference{Proceedings of the Workshop on Empowering Education with LLMs---the Next-Gen Interface and Content Generation held at AIED 2023, July 7, 2023, Tokyo, Japan.}

\title{Harnessing LLMs in Curricular Design: Using GPT-4 to Support Authoring of Learning Objectives}


\author[1]{Pragnya Sridhar}[%
email=pragnyas@andrew.cmu.edu,
]
\cormark[1]
\fnmark[1]
\address[1]{Language Technology Institute,
  Carnegie Mellon University, Pittsburgh, PA, USA}

\author[1]{Aidan Doyle}[%
email=adoyle@andrew.cmu.edu,
]
\cormark[1]
\fnmark[1]

\author[2]{Arav Agarwal}[%
email=arava@andrew.cmu.edu,
]
\address[2]{Computer Science Department, Carnegie Mellon University, Pittsburgh, PA, USA}

\author[2]{Christopher Bogart}[%
orcid={0000-0002-9421-8566},
email=cbogart@andrew.cmu.edu,
]

\author[2]{Jaromir Savelka}[%
orcid={0000-0002-3674-5456},
email=jsavelka@cs.cmu.edu,
]

\author[2]{Majd Sakr}[%
email=msakr@cs.cmu.edu,
]

\cortext[1]{Corresponding author.}
\fntext[1]{These authors contributed equally.}

\begin{abstract}
We evaluated the capability of a generative pre-trained transformer (GPT-4) to automatically generate high-quality learning objectives (LOs) in the context of a practically oriented university course on Artificial Intelligence. Discussions of opportunities (e.g., content generation, explanation) and risks (e.g., cheating) of this emerging technology in education have intensified, but to date there has not been a study of the models' capabilities in supporting the course design and authoring of LOs. LOs articulate the knowledge and skills learners are intended to acquire by engaging with a course. To be effective, LOs must focus on what students are intended to achieve, focus on specific cognitive processes, and be measurable. Thus, authoring high-quality LOs is a challenging and time consuming (i.e., expensive) effort. We evaluated 127 LOs that were automatically generated based on a carefully crafted prompt (detailed guidelines on high-quality LOs authoring) submitted to GPT-4 for conceptual modules and projects of an AI Practitioner course. We analyzed the generated LOs if they follow certain best practices such as beginning with action verbs from Bloom’s taxonomy in regards to the level of sophistication intended. Our analysis showed that the generated LOs are sensible, properly expressed (e.g., starting with an action verb), and that they largely operate at the appropriate level of Bloom's taxonomy, respecting the different nature of the conceptual modules (lower levels) and projects (higher levels). Our results can be leveraged by instructors and curricular designers wishing to take advantage of the state-of-the-art generative models to support their curricular and course design efforts.
\end{abstract}

\begin{keywords}
GPT-4 \sep
Large Language Models \sep
LLMs  \sep
Learning Objectives \sep
Automatic Generation \sep
Curricular Development  \sep
Course Design Automation \sep
Automated Content Generation
\end{keywords}

\maketitle

\section{Introduction}
Learning objectives (LOs) are the blueprints against which course content is designed. They provide instructors with a framework for content curation, instructional and assessment strategies, as well as enable learners to reflect on and plan their own learning of a course's knowledge and skills. Alignment between LOs, instructional strategies and assessments is a necessary prerequisite for internally consistent learning experience. When the three components are misaligned, learners may feel that tests do not assess what was covered in class, or instructors may notice that students earn a passing grade without mastering the material at the desired level. Hence, poor quality or missing LOs have a negative impact on the learning experience.\footnote{Eberly Center: Design \& Teach a Course. Available at: \url{https://www.cmu.edu/teaching/designteach/design/learningobjectives.html} [Accessed 2023-05-10]}

Creating effective LOs can be challenging and time-consuming for instructors, requiring substantial knowledge and experience in instructional design. Bloom \cite{bloom1956handbook,krathwohl2002revision,Adams2015BloomsTO} proposed a taxonomy organizing LOs into six levels (remember, understand, apply, analyze, evaluate, and create). This taxonomy helps educators articulate LOs that focus on concrete actions and behaviors, and target distinct levels of cognitive processes. For LOs to guide the selection of assessments, they must be measurable, i.e., it should be possible to evaluate whether learners attained the intended objective. Because of the complexity involved in authoring high-quality LOs, instructors often forego the task in lieu of more pressing duties such as authoring the course content or teaching.

Furthermore, instructors may have general notions of the learning objectives they want students to accomplish by the course's end. However, these notions may not meet the criteria of being well-defined and measurable learning objectives that focus on what a learner will achieve. To address this issue, our aim is to develop an approach that generates high-quality learning objectives and streamlines the objective-setting process. In our initial experiment, we investigate the potential of utilizing Language Models (LLMs) to generate effective learning objectives. This exploration serves as a stepping stone for future solutions that involve refining and augmenting the learning objectives initially provided by instructors.

Large Language Models (LLMs) are sophisticated AI models pre-trained on extensive textual data, that can generate human-level quality text. With appropriate prompting, an LLM might create high-quality LOs, alleviating the burden on the instructors. In this study we explore the potential of a state-of-the-art LLM (GPT-4) to support this task. We hypothesize that a well-prompted LLM can more efficiently produce candidate LOs, saving instructors' time.


To investigate the capability of GPT-4 to efficiently generate quality LOs in the context of a software development course on the practical integration of AI into applications, we analyzed the following research questions:

\begin{enumerate}
    \item[RQ1:] Are the generated LOs sensible, i.e., clear grammatically correct statements, addressing the relevant topic?
    \item[RQ2:] Do the LOs start with an appropriate action verb describing measurable behavior?
    \item[RQ3:] Do the conceptual module and project-related LOs target cognitive processes at the appropriate levels of Bloom's taxonomy? 
\end{enumerate}

\noindent To our best knowledge, this is the first study that proposes and evaluates automatic generation of LOs to drive the course design process, as opposed to generating LOs from an already existing course materials \cite{tran2018document}.

\section{Related Work}
Several works posit that LLMs could be an asset in pedagogy in several ways including Assessment Generation, Personalized Feedback System, generating lesson plans and asking questions about the best ways to teach a subject \cite{powerOfGPT,gpt4good,malinka2023educational}. While much work has already been done towards the applications of LLMs and course-content generation, LLMs have yet to be shown to generate course-guiding LOs. Thus, our focus on generating course-guiding LOs by leveraging the abilities of LLMs.

Tran et al. used IBM Watson to automatically generate LOs based on key-phrases extracted from course material documents paired with action verbs from Bloom's Taxonomy \cite{tran2018document}. In this work, we generate LOs using state-of-the-art LLM (GPT-4). While the system proposed in \cite{tran2018document} employs a similar definition of LOs to ours, the intended use is different. We generate LOs prior to course materials to potentially guide the content generation process. The system from \cite{tran2018document} assumes an existing collection of course material, and generate LOs from the collection.

Other work that harnesses LLMs to generate course material includes Multiple-Choice Question (MCQ) generation, such as the Question-Answer-Distractor pipeline in \cite{mcqT5}. Lu et al. utilized LLMs to effectively generate reading quizzes, confirming the effectiveness of the system in manual evaluation by 11 instructors across 7 different universities \cite{Lu2023ReadingQuizMakerAH}.

Adams comments on the direct application of Bloom's Taxonomy to developing LOs \cite{Adams2015BloomsTO}. By including action verbs associated with different levels of the taxonomy, educators are encouraged to think in terms of what students should be able to do at the end of the course. Additionally, the authors of \cite{tran2018document} generated the LOs by training a multi-layer perceptron (MLP) to learn which Bloom's Taxonomy action verb would best fit with each of the key phrases extracted from the collection of course material documents\cite{tran2018document}. In this work, the evaluation heavily focuses on verifying that the generated learning objectives begin with an action verb from an appropriate level of Bloom's Taxonomy.

In computing education context, LLMs have been shown to be highly effective at generating code and explanations of the code for entry-level programmers \cite{Sarsa_2022}. Such explanations have even been observed to out-class student explanations of the same code \cite{leinonen2023comparing}. Denny et al. discovered that well-structured prompts could yield correct solutions to many programming exercises \cite{denny2022conversing}; observations later confirmed by Savelka et al. \cite{savelka2023can,savelka2023thrilled}. Piccolo et al. demonstrated that LLMs can perform most entry-level programming tasks in the context of introductory bioinformatics course \cite{piccolo2023bioinformatics}. However, Savelka et al. \cite{savelka2023large} and Wermelinger \cite{wermelinger2023using} point to some limitations of LLMs in handling assessments from introductory programming classes. Phung et al. introduced a system that harnessed LLMs to provide precision feedback on syntax errors in students' code \cite{Phung2023GeneratingHF}.
Such feedback explanations went far beyond describing the code line-by-line. Sarsa et al. found such explanations were particularly valuable for student learning \cite{Sarsa_2022}. MacNeil et al. demonstrated that explanations of generated code can be offered at multiple different levels of abstraction \cite{MacNeill,MacNeil2}. 
In the near future, it is reasonable to expect LLMs to facilitate teacher-student exchanges similar to those that only occur in a classroom, and are invaluable to student learning \cite{tan2023applying}.

\section{Experiments}

\subsection{Model}
GPT has gained significant popularity due to its remarkable advancements in understanding and generating natural language text. It has demonstrated superior performance across various domains, including code generation \cite{gpt4code}, software engineering \cite{white2023chatgpt}, solving AI tasks \cite{shen2023hugginggpt}, and data augmentation \cite{dai2023auggpt}, showcasing its domain-invariant capabilities. In the context of education, a study conducted by Malinka et al. \cite{malinka2023educational} specifically investigated the impact of ChatGPT, a variant of the GPT model, on higher education with a focus on computer programming subjects. The authors provided evidence highlighting the effectiveness of ChatGPT in managing programming assignments, exams, and homework tasks.

Building on the success of its predecessors, GPT-4 represents a significant leap forward in language modeling technology \cite{openai2023gpt4}. Thus in our pursuit of generating course syllabus, we use the GPT-4 model (gpt-4). As of the writing of this paper, GPT-4 is by far the most advanced model released by OpenAI. The model is focused on dialog between a user and a system (i.e., an assistant).  


We set the \verb|temperature| of the model to 0.7, which is the default. The higher the \verb|temperature| the more creative the output but it can also be less factual. As the temperature approaches 0.0, the model becomes deterministic and can be repetitive. We set \verb|max_tokens| to 2,000 tokens (a token roughly corresponds to a word). This parameter controls the maximum length of the completion (i.e., the output). Note that GPT-4 has an overall token length limit of 8,192 tokens, comprising both the prompt and the completion.\footnote{There is also a variant of the model that supports up to 32,768 tokens.} We set \verb|top_p| to 1 (default). This parameter is related to \verb|temperature| and also influences creativeness of the output. We set \verb|frequency_penalty| to 0, which allows repetition by ensuring no penalty is applied to repetitions. Finally, we set \verb|presence_penalty| to 0, ensuring no penalty is applied to tokens appearing multiple times in the output.

\subsection{Experimental Design}
\label{sec:experimental_design}

\setcounter{figure}{1}
\begin{figure}[t]
\footnotesize
\begin{Verbatim}[frame=single,commandchars=\\\{\}]
You are a curricular development expert system focused on authoring LOs. Learning 
objectives are brief, clear statements that describe the desired learning outcomes 
of instruction. \textcolor{gray}{[601 characters ...]} LOs should use action verbs. LOs should be 
measurable.

A well-constructed learning objective contains three parts \textcolor{gray}{[392 characters ...]}

1. BEHAVIOR
The behavior is the real work to be accomplished by the student specified by an 
action verb that connotes observable and measurable behaviors. 
\textcolor{gray}{[2,497 characters ...]}

2. CONDITIONS
This is a statement that describes the exact conditions under which the defined 
behavior is to be performed. \textcolor{gray}{[117 characters ...]}

3. DEGREE
This is a statement that specifies how well the student must perform the behavior 
\textcolor{gray}{[171 characters ...]}

Conceptual LOs are focused on students' knowledge and understanding (i.e., the first 
two levels of Bloom's taxonomy).

\textcolor{gray}{[18 example LOs (1,540 characters)  ...]}

Project LOs are focused on students' skills and behaviors (i.e., the higher levels 
of Bloom's taxonomy).

\textcolor{gray}{[12 example LOs (1,261 characters)  ...]}

Here are some criteria to satisfy in order to create an effective learning 
objective:
1. LOs should be student-centered. \textcolor{gray}{[114 characters ...]}
2. LOs should focus on specific cognitive processes. \textcolor{gray}{[530 characters ...]}
3. LOs should use action verbs.
4. LOs should be measurable \textcolor{gray}{[105 characters ...]}

The user will provide you with the name of the course, brief description of the
course goals, the name of the module, and the type of the LOs to be developed. Based 
on these you respond with a list of well-designed effective LOs (5-10 items).
\end{Verbatim}
\caption{The system prompt. The figure shows the essential elements of the system prompt guiding the model towards generating high-quality LOs. The grey annotations are used as replacement for extensive parts of the prompt that could not be fit into the figure.}
\label{fig:system_prompt}
\end{figure}

\noindent To generate LOs, we utilize the system prompt illustrated in Figure \ref{fig:system_prompt}. The system prompt guides the GPT-4 model towards the desired behavior. The prompt contains brief guidelines on how LOs should be structured and what properties are desirable. These guidelines were informed by various university materials on course design.\footnote{Eberly Center: Design \& Teach a Course. Available at: \url{https://www.cmu.edu/teaching/designteach/design/learningobjectives.html} [Accessed 2023-05-10]; Center for Excellence in Teaching and Learning at UCONN: Developing LOs. Available at: \url{https://cetl.uconn.edu/resources/design-your-course/developing-learning-objectives/} [Accessed 2023-05-10]} The guidelines instruct the system to generate LOs for conceptual modules as well as projects. The LOs should start with an action verb describing the behavior, state the conditions under which it is to be performed, and the degree of mastery the learners should attend. The prompt also provides many example LOs. These can be related to conceptual modules that focus on the two lower levels of Bloom's taxonomy, e.g.:

\begin{quote}
    Define DevOps from organizational, cultural and technical perspectives.
\end{quote}

\noindent The LOs from projects focus on behaviors described by action verbs from higher levels of Bloom's taxonomy, e.g.:

\begin{quote}
    Design and implement Continuous Integration and Continuous Delivery for a Node.JS application.
\end{quote}

\noindent The prompt emphasizes the difference between the conceptual module and project related LOs. Hence, given the same topic the LOs for the conceptual module are expected to differ substantially from those generated for the project in terms of their focus on different levels of Bloom's taxonomy.

\begin{figure}[t]
\footnotesize
\begin{Verbatim}[frame=single,commandchars=\\\{\}]
COURSE NAME: AI Practitioner
COURSE GOALS: In this course, learners gain hands-on experience solving real-world 
problems by completing projects focused on developing AI/ML-enabled systems. It is 
our goal that students will develop the skills needed to become sophisticated
developers of AI/ML-based systems. Specifically, students are exposed to real-world 
data and scenarios to learn how to:
- Integrated different types of AI/ML systems into their applications, recognize 
their capabilities and limitations.
- Explain the effects of data quality, quantity, and representativeness on the
performance of AI/ML systems.
- Inspect, validate, and critically assess outputs of AI/ML systems.
- Use AI/ML components via cloud APIs or locally run libraries from different areas, 
such as language technologies, or computer vision, including state-of-the art 
generative models.
- Discuss the advantages and disadvantages of different computing devices and 
environments for deployment of Artificial Intelligence systems.
Through this process, we aspire for our students to become sophisticated, 
independent, and resilient problem solvers who are able to overcome challenges and 
learn.
MODULE NAME: \textcolor{orange}{\string{\string{module\string}\string}}
LOs TYPE: \textcolor{orange}{\string{\string{module_type\string}\string}}
EXPECTED OUTPUT:
1. Text of LO1.
2. Text of LO2.
...
\end{Verbatim}
\caption{The User Message. The figure shows the user message template specifying the context for the LOs to be generated. The grey orange tokens are replaced with module specific information.}
\label{fig:user_message}
\end{figure}

The context of the particular course being designed, i.e., AI Practitioner, is provided to GPT-4 via the user message. The full template of the user message used in this work is shown in Figure \ref{fig:user_message}. It provides the name of the course, brief description of the high-level course goals, placeholders for a module name (e.g., ``Generative Models'' or ``AI/ML in the Cloud'') and module type (i.e., either a ``conceptual module'' or a ``project''). To generate LOs for each conceptual module and each project, a separate message is used with the placeholders filled in accordingly. The dynamically constructed prompts, i.e., the system prompt and the user message, are submitted individually to OpenAI's GPT-4 API using the \verb|openai| Python library.\footnote{GitHub: OpenAI Python Library. Available at: \url{https://github.com/openai/openai-python} [Accessed 2023-05-10]}

We extracted the generated LOs from the GPT-4 responses and analyzed them to answer the three research questions. To answer RQ2, we used a simple regular expression to extract the action verb from each learning objective. To answer RQ3, we evaluated the generated LOs using both automatic methods and human annotation. We presented the 127 of the LOs to 3 graduate computer science students and asked them to classify them into the individual levels of Bloom's Taxonomy. Out of these, 101 LOs were annotated by all three of the annotators. We also automatically classified the generated LOs into the individual levels of Bloom's taxonomy using the approach mentioned in \cite{2022.EDM-short-papers.55}. They trained a binary classifier for each Bloom's Taxonomy category using a dataset of 21,380 LOs from 5,558 university courses. We used the same models to predict the Bloom's Taxonomy level of the generated LOs.  

\subsection{Results}



\begin{figure}
  \centering

  \begin{minipage}{0.32\textwidth}
    \centering
    \includegraphics[width=\linewidth]{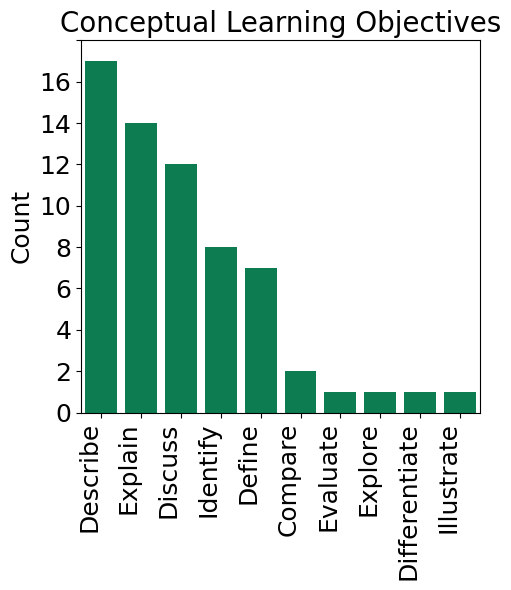}
    \label{fig:concept-action-verb}
  \end{minipage}\hfill
  \begin{minipage}{0.65\linewidth}
    \centering
    \includegraphics[width=\textwidth]{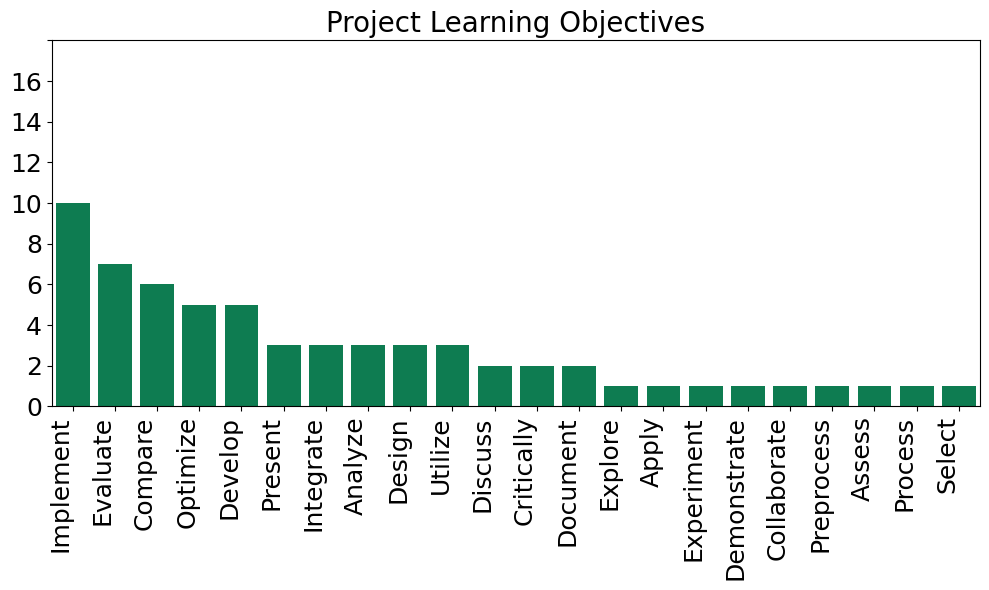}
    \label{fig:project-action-verb}
  \end{minipage}

  \caption{Distribution of action verbs across module types.}
  \label{fig:action-verb}
\end{figure}

Figure \ref{fig:action-verb} shows that LOs for conceptual modules heavily employ a small number of action verbs such as ``describe,'' ``discuss,'' ``explain,'' ``identify'' and ``define''. This is expected since these verbs are geared toward conceptual learning. The LOs for projects use more diverse set of action verbs. The examples include ``implement,'' ``optimize,'' ``develop,'' or ``utilize.'' These verbs also appear sensible as they focus on activities and skills as opposed to conceptual knowledge.

\begin{figure}

  \centering
\includegraphics[width=.8\linewidth]{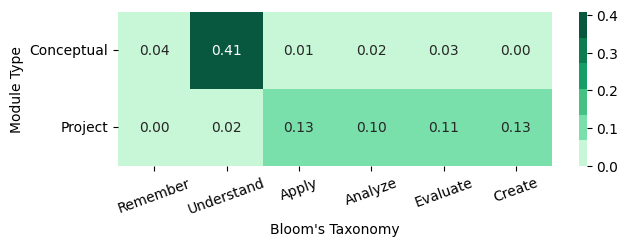}
  \caption{Bloom's Taxonomy Categorization of Generated LOs using BERT}
  \label{fig:bloom-heatmap}
\end{figure}

\begin{figure}

  \centering
\includegraphics[width=.8\linewidth]{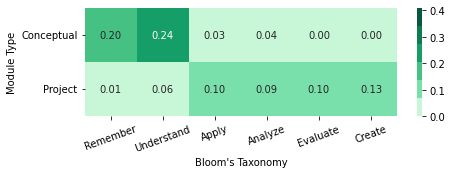}
  \caption{Bloom's Taxonomy Categorization of Generated LOs based on Human Annotation}
  \label{fig:bloom-heatmap-human}
\end{figure}

Figure \ref{fig:bloom-heatmap} reports the distribution of LOs across the conceptual modules and projects categorized in terms of Bloom's Taxonomy categories, using the classifier proposed in \cite{2022.EDM-short-papers.55}. The LOs generated for conceptual modules predominantly fall under the ``Understand'' level, while the LOs generated for projects are spread across the ``Apply,'' ``Analyze,'' ``Evaluate,'' and ``Create'' levels. This is expected. Figure \ref{fig:bloom-heatmap-human} shows similar distribution where instead of the BERT-based classifier we use levels of Bloom's taxonomy assigned by each of our human annotators. Note that the counts in this figure are normalized, as each LO was annotated three times during human annotation. The distribution differs a bit but we can still observe the LOs for the conceptual modules mostly employing the action verbs from the lower levels of Bloom's taxonomy, whereas the LOs for projects use action verbs from the higher levels. 

\section{Discussion}

\subsection{RQ1:  Are the generated LOs sensible?}

Overall, the LOs are largely sensible. They describe key sub-concepts related to the relevant topic, and mostly focus on one or two separate cognitive processes, e.g.:

\begin{quote}
Explain the key concepts and techniques used in computer vision, such as image processing, feature extraction, and object recognition.
\end{quote}

\noindent GPT-4 generates LOs with action verbs such as ``describe,'' ``explain,'' and ``discuss'' for conceptual modules, and ``implement,'' ``evaluate'' and ``develop'' for project-based materials. 
While the LOs are sensible they sometimes lack specific focus. For example, one such generated LO was ``Implement a basic AI/ML model using Python libraries to solve a simple classification or regression problem.''  While this is measurable, the term ``Python libraries'' is too broad, and would benefit from being more focused (e.g., ``scikit-learn''). This issue may be solvable with further prompt tuning. 

\subsection{RQ2:  Do the LOs start with an appropriate action verb?}
All the LOs start with action verbs. The distribution of action verbs across LOs for conceptual modules and projects is as expected. Action verbs such as ``describe'' and ``explain'' should be associated with conceptual materials often focused on declarative knowledge. Whereas action verbs such as ``implement'' and ``develop'' should be associated with projects geared towards procedural knowledge.

We found that some generated LOs (Figure \ref{fig:action-verb}) employed action verbs that were not included in the extensive list of example verbs provided to the model via the prompt. Specifically, these were ``optimize,'' ``preprocess,'' ``explore,'' ``document,'' ``implement,'' ``utilize,'' and ``process''. Of these, only ``utilize'' and ``implement''  appeared in the examples mentioned in the prompt. Of the 127, there were 26 LOs starting with action verbs not included in the list of examples: 25 of these were for projects and one was for a conceptual module. Moreover, there were 13 LOs with action verbs that have that were not provided anywhere in the prompt. 




Furthermore, there are 11 generated LOs that utilize multiple action verbs, e.g., ``Evaluate the performance of computer vision models using appropriate metrics and develop strategies to improve their accuracy and reliability.'' These could have been separate LOs (e.g., ``Evaluate the performance of computer vision models using appropriate metrics.'' and ``Develop strategies to improve the accuracy and reliability of computer vision models'').

\subsection{RQ3: Do the LOs target cognitive processes at the appropriate levels of Bloom's taxonomy? }

The results of applying BERT classifier as described in Section \ref{sec:experimental_design} show that the GPT-4 model generates LOs that largely operate on the expected levels of Bloom's Taxonomy (Figure \ref{fig:bloom-heatmap}). Conceptual modules are more focused on declarative knowledge, and the LOs mostly employ action verbs from the lower two levels of Bloom's taxonomy (Remember and Understand). Projects are focused on procedural knowledge, and the LOs mostly use action verbs from the higher four levels (Apply, Analyze, Evaluate and Create). Based on the BERT classifier, it appears that the generated LOs are geared towards the appropriate types of cognitive processes.

The classifier proposed in \cite{2022.EDM-short-papers.55} may have some limitations. We noticed that two of the generated LOs were not assigned to any of the Bloom's Taxonomy levels, while five LOs were assigned multiple categories. Note that these issues pertain to a relatively small proportion of the generated LOs. 

We presented the generated LOs to human annotators to validate the BERT-based classification. As seen in Figure \ref{fig:bloom-heatmap}, most LOs generated for conceptual modules were classified as targeting the Understand and Remember levels of Bloom's taxonomy, and  most project LOs were classified as \emph{not} Understand or Remember. This same distribution can be observed in \ref{fig:bloom-heatmap-human}, although with a higher frequency for humans classifying conceptual LOs as 'Remember' instead of 'Understand'.  When conducting the human annotation, we used Cohen's $K$ to examine our inter-rater agreement. The average agreement among raters for classifying LOs into each of the six individual levels of Bloom's Taxonomy was 0.31, corresponding to fair agreement. By mapping the BERT and human classifications to the corresponding LO categories, we were able to observe an agreement between the majority-vote annotation and the BERT classification of 0.62, demonstrating substantial agreement between the human and BERT classification of LOs into either the 'Remember' and 'Understand', or the 'Apply', 'Analyze', 'Evaluate', and 'Create' levels of Bloom's Taxonomy. 

\section{Implications for Education Practice}

Automatic generation of LOs could significantly ease the workload of educators, allowing them to focus more on teaching and student interaction. Automation could improve the quality of LOs. Reducing the cost of authoring LOs could open up a possibility of personalized LOs for each student based on their individual strengths, weaknesses, and progress. On the other hand, the over-reliance on automated systems could potentially lead to a loss of pedagogical nuance and adaptability. The standardization might stifle creativity and innovation in teaching methods. Therefore, integrating such systems into teaching practice should be handled with caution, ensuring they serve as a supportive tool rather than a replacement for educators' expertise.

\section{Limitations}
LOs drive the entire course development process. Mistakes in authoring LOs may cascade and snowball into larger issues manifesting in low-quality course content. LLMs are relatively new technology, and there may be skepticism or resistance from educators and experts regarding the reliability and validity of using LLMs to generate LOs. Addressing these concerns and gaining acceptance is essential for potential widespread adoption of the proposed approach. Additionally, there is a need for human validation of the generated LOs. While LLMs can assist in the initial generation process, human expertise and judgment are crucial to ensure the accuracy, relevance, and appropriateness of the generated LOs. Some instructors or institutions have suggested use of GPT to be unethical because of its training on copyrighted materials. Hence, its products may not be usable in institutions with policies preventing such use.


\section{Conclusions and Future Work}
This paper explored the use of LLMs for generating LOs in the context of practically oriented university course on AI. Prior work demonstrated the effectiveness of LLMs in various tasks in educational context, even in generating various elements of course content. This work highlights the potential of LLMs for generating LOs to support curricular development. We evaluated the effectiveness of GPT-4 on this task. We found that the generated LOs are sensible, properly expressed (e.g., starting with an action verb), and that they largely operate at the appropriate level of Bloom's taxonomy respecting the different nature of the conceptual modules (lower levels) and projects (higher levels). These findings can be leveraged by instructors and curricular designers wishing to take advantage of the state-of-the-art generative models to support their curricular and course design efforts. In future work, we plan to further evaluate the generated LOs, especially in terms of the LOs being measurable. This may include generating LOs for existing courses with existing human-created LOs in order to compare the two. In addition, GPT-4 could also be used for developemn of assessment strategies for the generated LOs.

\bibliography{sample-ceur}




\end{document}